\documentclass{elsarticle}

\usepackage{todonotes}
\usepackage{fancybox}

\usepackage{amsmath,amssymb,amsfonts}
\usepackage{algorithmic}
\usepackage{graphicx}
\usepackage{textcomp}
\usepackage{xcolor}
\usepackage{times}
\usepackage{url}
\usepackage{latexsym}
\usepackage{multirow}
\usepackage{placeins}
\usepackage{pdfpages}
\usepackage{amssymb}
\usepackage{natbib}
\usepackage{rotating}
\usepackage{todonotes}
\usepackage{wrapfig}
\usepackage[colorlinks=false]{hyperref} 
\usepackage{lipsum,booktabs}
\usepackage{multicol}
\usepackage{amsmath}
\usepackage{cleveref}
\usepackage{caption}
\usepackage{subfigure}
\usepackage{multirow}
\usepackage{lineno,hyperref}

\journal{Applied Soft Computing}

\def\BibTeX{{\rm B\kern-.05em{\sc i\kern-.025em b}\kern-.08em
    T\kern-.1667em\lower.7ex\hbox{E}\kern-.125emX}}
\newcommand{\CTF}{\texttt{CTF}}

\makeatletter
\def\ps@pprintTitle{%
  \let\@oddhead\@empty
  \let\@evenhead\@empty
  \let\@oddfoot\@empty
  \let\@evenfoot\@oddfoot
}
\makeatother

\makeatletter
\def\blfootnote{\xdef\@thefnmark{}\@footnotetext}
\makeatother

\begin{document}

\begin{frontmatter}

\title{Cross-SEAN: A Cross-Stitch Semi-Supervised  Neural Attention Model for COVID-19 Fake News Detection}

\author{
\textbf{William Scott Paka\textsuperscript{$1$}\textsuperscript{$\dagger$}\textsuperscript{$\star$}\quad Rachit Bansal\textsuperscript{$2$}\textsuperscript{$\star$}\textsuperscript{$\ddagger$}\fnref{ec}\quad Abhay Kaushik\textsuperscript{$3$}\textsuperscript{$\ddagger$}\quad \\ Shubhashis Sengupta\textsuperscript{$4$}\quad Tanmoy Chakraborty\textsuperscript{$1$}}\\
\textsuperscript{$1$}IIIT-Delhi, India\quad \textsuperscript{$2$}DTU-Delhi, India\quad \textsuperscript{$3$}IIT-Kanpur, India\quad \textsuperscript{$4$}Accenture Labs, India\\
\texttt{\{william18026, tanmoy\}@iiitd.ac.in\\rachitbansal\_2k18ee152@dtu.ac.in; kabhay@iitk.ac.in\\shubhashis.sengupta@accenture.com}}

\begin{abstract}
As the COVID-19 pandemic sweeps across the world, it has been accompanied by a tsunami of fake news and misinformation on social media. At the time when reliable information is vital for public health and safety, COVID-19 related fake news has been spreading even faster than the facts. During times such as the COVID-19 pandemic, fake news can not only cause intellectual confusion but can also place people's lives at risk. This calls for an immediate need to contain the spread of such misinformation on social media.

\textcolor{black}{We introduce \textbf{\CTF}, a large-scale COVID-19 Twitter dataset with labelled genuine and fake tweets.}
\textcolor{black}{Additionally, we propose {\bf Cross-SEAN}, a cross-stitch based semi-supervised end-to-end neural attention model which leverages the large amount of unlabelled data. Cross-SEAN partially generalises to emerging fake news as it learns from relevant external knowledge.} We compare Cross-SEAN with seven state-of-the-art fake news detection methods. We observe that it achieves $0.95$ F1 Score on \CTF, outperforming the best baseline by $9\%$.
We also develop {\bf Chrome-SEAN}, a Cross-SEAN based chrome extension for real-time detection of fake tweets.
\end{abstract}

\end{frontmatter}

\section{Introduction \label{sec:intro}}
\blfootnote{\textsuperscript{$\dagger$}Corresponding author.}
\blfootnote{\textsuperscript{$\star$}Equal contribution.}
\blfootnote{\textsuperscript{$\ddagger$}Work done during an internship at IIIT-Delhi.}

The increase in accessibility to Internet has dramatically changed the way we communicate and share ideas. Social media consumption is one of the most popular activities online. Nowadays, it is a trend to rely on such platforms for news updates. The absence of a verification barrier allows misinformation on sites online. \textcolor{black}{Due to the complexity of the issue, the definition of “fake news” is not well defined. A few definitions used in prior studies are as follows: `A news article that is intentionally and verifiably false' \cite{allcott2017social, shu2017fake} relating to news that are deceptive in nature, `A news article or message published and propagated through media, carrying false information regardless of the means and motives behind it' relating to various forms of false news and misinformation \cite{kshetri2017economics, kucharski2016study, golbeck2018fake,varshney2020review,vishwakarma2020recent}. A few broader definitions by Zhou et al. \cite{zhou2020survey} state, `Fake news is false news', `Fake news is intentionally false news published by a news outlet.' For our purpose, we define COVID-19 fake tweet as any tweet with information which contradicts the statements released by the governmental health organisations\footnote{\url{https://en.wikipedia.org/wiki/List\_of\_health\_departments\_and\_ministries}}, and genuine tweets to be the tweets obtained from their official accounts.}

On 30 January 2020, \textcolor{black}{The World Health Organisation (WHO)} has declared COVID-19 to be a Public Health Emergency of International Concern and issued a set of Temporary Recommendations. 
A recent study observed 25\% increase in average user social media activity due to the global lockdown \cite{yang2020prevalence}. UNESCO stated, ``during this coronavirus pandemic, fake news is putting lives at risk.'' Fake news, ranging from the speculations around origin of the virus to baseless prevention and cures, is spreading rapidly without any valid evidence. WHO has recently declared the spread of COVID-19 related misinformation as an `Infodemic'; according to their definition, ``An infodemic is an overabundance of information, both online and offline. It includes deliberate attempts to disseminate wrong information to undermine the public health response and advance alternative agendas of groups or individuals.'' WHO, CDC (Centers for Disease Control and Prevention) and other other government bodies have set up specific web pages in order to curb major misconceptions about the virus and to maintain public awareness. Any single false news that gains enormous traction can negate the significance of a body of verified facts. When a tweet with misinformation is retweeted by an influential person or by a verified account, the marginal impact grows largely. The analysis, identification, and elimination of fake news thus have become a task of utmost importance. Therefore, there is an immediate need to detect the fake news and stop their spreading.

Till now, no verification barrier exists that can authenticate the content being shared on social media platforms. Due to this, quite often, general people are misinformed when an unreliable news or information is shared irrespective of intentions. With increase in reliance on social media platforms such as Twitter and Facebook for information, the spread of misinformation also tends to increase. Fake news is usually targeted for financial or political gain with click-bait titles or advertisement links gaining user attention. The spread of fake news is proven to be a threat in the past during global events such as US 2016 elections and the Brexit. Studies showed that automated bots are used for spreading fake content \cite{howard2017junk}; however, all the posts of bots cannot be considered as fake since they are devised to post non-fake content too. Genuine users seldom fall prey to fake content, and with uninformed knowledge sharing among their network makes genuine users major contributors to its spread.

Twitter is one of the largest micro-blogging platforms with over 1.5M daily active users combating fake news since a long time. The major exploitation of fake news is highlighted during the 2016 U.S. presidential election campaign. The existence of `echo chamber effect' on social media allows biased information to be spread wider and deeper \cite{jamieson2008echo}. Tweets containing fake content show far wider reach, spreading rapidly than normal tweets, and such variations in propagation can be clearly observed in tweets related to political news. Such tweet propagation behaviour is partly due to the innate nature of users to retweet content which is provocative, aligning to their beliefs, irrespective of the truthfulness of the content. Social and psychological factors with `valence effect' \cite{harvey2018new} play an important role in the spread of fake news. Studies also showed the involvement of bots to create and spread fake news \cite{howard2017junk}. News involving any political figure in power create huge fluctuations in stock markets and trades economically. For example, a 2013 tweet `Breaking: Two Explosions in the White House and Barack Obama is injured', from a hacked Associated Press account created a loss of \$136 billion worth of stock value \cite{fisher2013syrian}. Twitter has a long history of accounts getting stolen, and hackers with motivations to create mass hysteria take control of verified accounts for wider spread of hoax. Although the character limit helps the amount of textual content being shared, other forms of content such as images, videos and links are also exploited to spread false information. Twitter usually deletes tweets and users that are flagged post-verification; however, this is not a scalable solution for automated fake news verification.

\begin{figure}
    \centering
    \includegraphics[width=\linewidth]{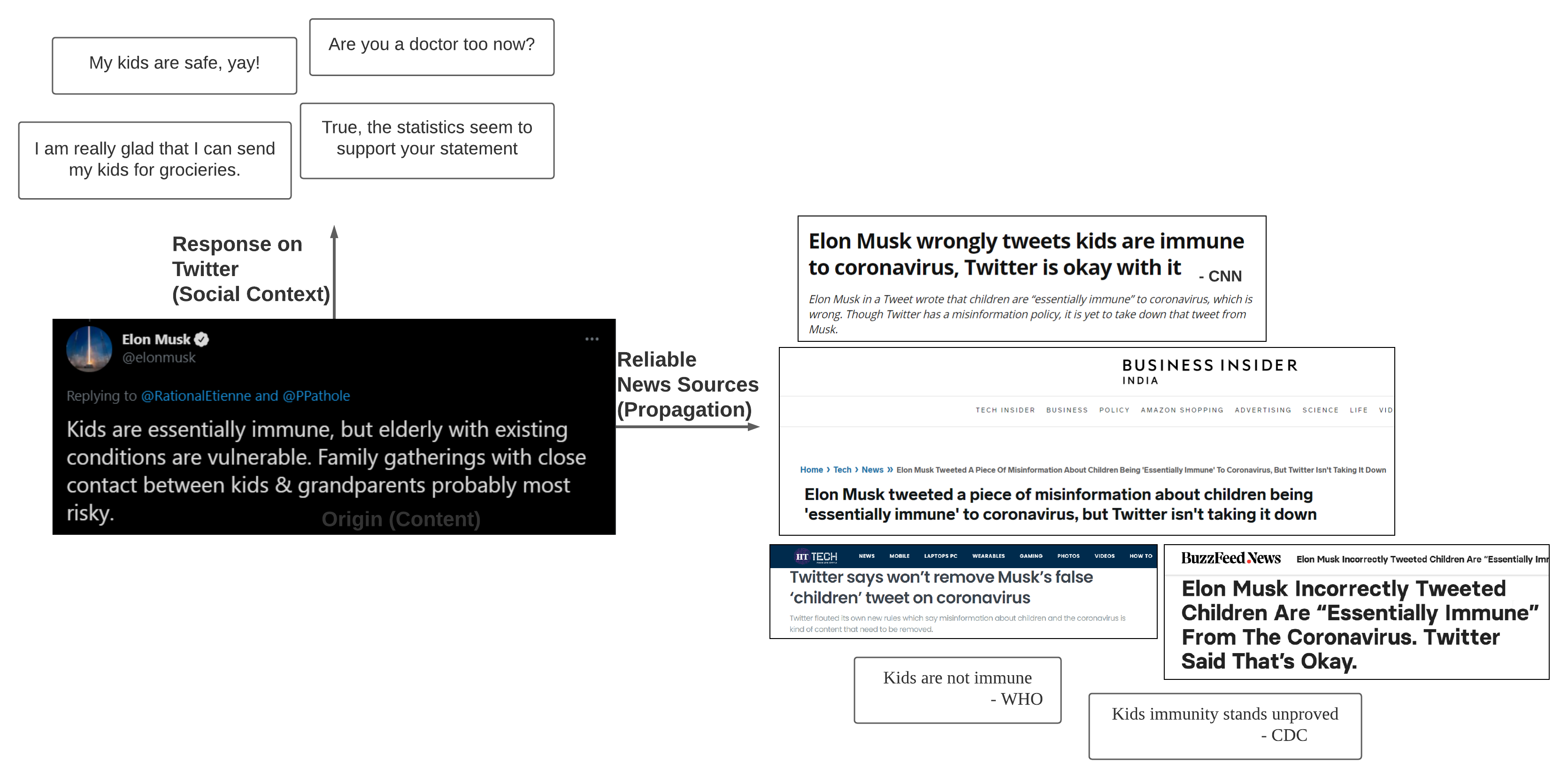}
    \caption{\textcolor{black}{An example of {\em origin}, {\em propagation} and {\em social context} of a popular misinformation. The responses for a tweet with misinformation seem to be coherent to it, and could ultimately spread it wider and deeper into the follower networks. Both the tweets and responses contradict the reliable news source.}}
    \label{fig:introdiag}
\end{figure}

\begin{figure}[!t]
  \centering
  \subfigure[Friend count vs follower count for users tweeting genuine and fake tweets (plotted across 500 samples for each class).]{\includegraphics[trim=0cm 0cm 0cm 0cm, clip = true, width=0.48\linewidth]{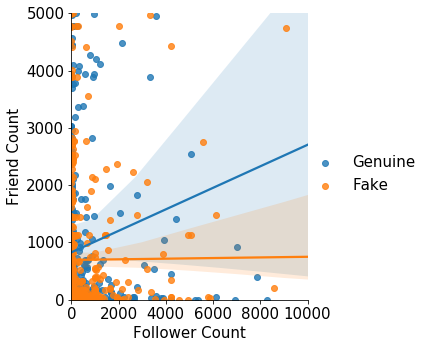}\label{fig:counts_fake_genuine_a}}\quad
  \subfigure[Favourite count vs Retweet count of users posting genuine and fake tweets (plotted across 250 samples for each class).]{\includegraphics[trim=0cm 0cm 0cm 0cm, clip = true, width=0.48\linewidth]{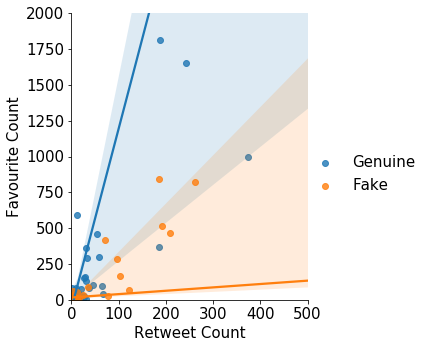}\label{fig:counts_fake_genuine_b}}\quad
  \caption{Correlations between (a) user features and (b) tweet features for genuine and fake tweets. In (b), note that a large number of samples are present close to the origin.}
  \label{fig:counts_fake_genuine}
\end{figure}

Due to the lockdown and work from home conditions during COVID-19 pandemic, Twitter witnessed a 30\% rise in daily average usage. With isolation from the external world, users turn to social media platforms for any updates related to the pandemic. Due to uninformed knowledge, users tend to retweet content which may not be totally accurate. At the beginning of the pandemic, very limited information is available to the public on the realities of the virus. Even verified users such as Elon Musk tweeted stating that ``Kids are essentially immune'' which provides statistical evidence in which there are no infected people below the age 19. Public health experts later released a statement debunking his claim. \textcolor{black}{We illustrate this in Fig. \ref{fig:introdiag}, showing the tweet with misinformation by Elon Musk along with the `responses received in Twitter' and `reliable news sources statements'. We can notice that the retweets are coherent to the misinformed tweet which spread the misinformation across other networks, wider and deeper. The news from verified sources state otherwise, clearly debunking the said statement} Due to the scarcity of reliable information source, multiple fact checking sites depend on statements released by Public Health bodies. Although few users tweet and retweet false content without any ill-intention, there exist users who create and spread false news for political gains. 
Diffusion of fake tweets and genuine tweets vary in a pandemic setting such as this \cite{masud2020hate}. Tweeting a political tweet with false information multiple times from several accounts with various trending hashtags, called `Hashtag hijacking' is also observed. Fig. \ref{fig:counts_fake_genuine_a} shows the count of favourites and retweets for both genuine and fake tweets, whereas Fig. \ref{fig:counts_fake_genuine_b} shows the friends and followers count of users posting genuine and fake tweets. We can clearly observe from Fig. \ref{fig:counts_fake_genuine_a} that genuine tweets tend to have higher favourite count compared to retweet count whereas the fake tweets tend to have higher retweet count, propagating the false information to a wider range. We can also observe from Fig. \ref{fig:counts_fake_genuine_b} that users posting genuine content have higher number of friends than followers, and users posting fake content have higher number of followers than friends -- this setting again allows spread of fake news towards larger audiences through the users posting fake content.

\textcolor{black}{The rest of our paper is organised as follows. We discuss related works on fake news detection and semi-supervised models for text classification in Section \ref{sec:rel}. Section \ref{sec:dataset} describes our four-stage dataset collection and annotation processes, which is followed by further analysis of the dataset on various  aspects in Section \ref{sec:analy}. The proposed Cross-SEAN model  and training strategies are introduced in Section \ref{sec:model}, while its evaluation and a detailed ablation study are shown in Section \ref{sec:experiment}. For real-time usage of Cross-SEAN, the developed chrome extension, Chrome-SEAN and the user study are described in Section \ref{sec:chrome}. Finally, the paper is concluded with discussions, shortcomings and future work in Section \ref{sec:concl}.}

{\bf Our contribution I: \CTF - A COVID-19 fake news dataset and its analysis.} With the aforementioned concerns, it is evident that more research is required to detect and neutralise fake tweets and keep the users warned. Although research communities are interested to work on the challenging task of COVID-19 fake news detection which is one of the pressing issues of our time, the absence of a publicly available labelled COVID-19 misinformation dataset is a major bottleneck to design automated detection models.
Also, not everyone possesses the resources to collect such a dataset, as it is cumbersome. We fill this gap by introducing \CTF, the first COVID-19 Twitter fake news dataset, consisting of a mixture of both labelled and unlabelled tweets. Our dataset contains a total of $45.26K$ labelled tweets, among which $18.55K$ are labelled as `genuine' and $26.71K$ as `fake'. In addition, it contains $21.85M$ unlabelled tweets, which can be used to enrich the diversity of the dataset, in terms of linguistic and contextual features in general. A detailed analysis of the dataset unfolds many interesting observations. E.g., fake news content tends to -- (i) accompany less URLs and more multimedia content, (ii) receive much lesser likes and retweets, (iii) exhibit mostly neutral and negative sentiment, as compared to genuine content. Our dataset collection is a four stage process, starting from hydration of Tweets, collection of supporting statements, usage of fine-trained Transformer models such as BERT and RoBERTa, to manual annotation.
As COVID-19 is an emerging topic, we rely on certain government health organisations and fact checking sites such as PolitiFact, Snopes, TruthOrFiction, etc, which release statements on widely popular misconceptions. We then use tweets on the collected facts using BERT and RoBERTa to identify supporting or contradicting claims, which are then partially annotated. The major part of our genuine tweets are taken from governmental health organisations.

{\bf Our contribution II: Cross-SEAN.} Two major issues in any fake news detection task are the lack of labelled data to train a deep neural model and the inability to detect fake news that are different from the training data (emerging fake news). To address these issues, 
we propose Cross-SEAN, a cross-stitch based semi-supervised attention neural model. Cross-SEAN works in a semi-supervised way leveraging the vast unlabelled data to learn the writing style of tweets in general. It considers user metadata, tweet metadata, and external knowledge in addition to tweet text as its inputs. External knowledge is collected on the fly in the form of stances close to tweets from trusted domains and allows a way for Cross-SEAN to not restrict to the train data, as external knowledge can contain information which is absent in the train data partially helping with early detection. When multiple inputs are involved, simple concatenation of layers might undermine few inputs' significance on the model. We employ cross-stitch mechanism which provides a way to find the optimal combination of model parameters that are used to pass the inputs to various sections of the network. 
Attention mechanisms have the ability of `attending to' particular parts of the input when processing the data, allowing Cross-SEAN to be capable of representing the words which are being concentrated on, for a given tweet text. 


We compare Cross-SEAN with seven state-of-the-art models for fake news detection. 
Experimental results show that Cross-SEAN achieves $0.95$ F1 Score on \CTF, outperforming seven baselines by at least $9\%$. We show comparative evaluation of baselines with Cross-SEAN on various features and present a thorough ablation study of Cross-SEAN to understand the importance of different features and various components of the objective function.  

{\bf Our contribution III: Chrome-SEAN.} For easy and real-time usage by Twitter users, we finally introduce a chrome extension, called Chrome-SEAN which uses Cross-SEAN to classify a tweet while in the tweet page. To evaluate Chrome-SEAN, we collect feedback from human subjects. We further perform online learning conditioned on the feedback and the confidence of model. The extension is deployed and configured to handle concurrent requests.

In summary, our major contributions are four-fold:
\begin{itemize}
    \item  \textbf{\CTF}, the first labelled COVID-19 misinformation dataset.
    \item {\bf Cross-SEAN}, a model to curb COVID-19 fake news on Twitter. It is one of the few semi-supervised models introduced for the task of fake news detection.
    \item Detailed analyses of the dataset to unfold the underlying patterns of the COVID-19 related fake tweets.
    \item  {\bf Chrome-SEAN}, a chrome extension to flag COVID-19 fake news on Twitter.
\end{itemize}

\vspace{4mm}
\noindent\shadowbox{\begin{minipage}[t]{.95\columnwidth}
\noindent{\bf Reproducibility:} We have made the code and the \CTF\ dataset public at \url{https://github.com/williamscott701/Cross-SEAN}.  Section \ref{sec:experiment} describes more about the settings to reproduce the results.
\end{minipage}}

\section{\color{black}{Related Work} \label{sec:rel}}

As our work revolves around fake news and semi-supervised learning, we present the related work in two parts: (i) fake news detection, and (ii) text-based semi-supervised learning. Due to the abundance of literature in both these areas, we focus our attention to those studies which we deem as pertinent to the current work.

\textbf{Fake news detection:}
Fake news or misinformation on social media has gained a lot attention due to the exponential usage of social media. Some of early studies tried to detect fake news on the basis of linguistic features of text \cite{castillo2011information, qazvinian2011rumor, gupta2014tweetcred}. A group of recent approaches have used temporal linguistic features with recurrent neural network (RNN) \cite{ma2016detecting} and modified RNN \cite{chen2018call, SAHOO2021106983} to detect fake news. Hybrid approaches by Kwon et al. \cite{kwon2017rumor} combined user, linguistic, structural and temporal features for fake news classification. Lately, convolution networks have been adopted along with recurrent networks to detect fake news \cite{liu2018early, KALIYAR202032}. \textcolor{black}{Malhotra and Vishwakarma \cite{malhotra2020classification} used the graphical convolutional networks and transformer-based encodings for the task of rumor detection of tweets. They leveraged the structural and graphical properties of a tweet's propagation and tweet's text}. Since satire can also lead to spread of misinformation, Rubin et al. \cite{rubin2016fake} proposed a classification model using 5 features to identify satire and humour news. Another study focused on detecting fake news using n-gram analysis through the lenses of different feature extraction methods \cite{ahmed2018detecting}. Granik and Mesyura \cite{granik2017fake} detected fake news using Naive Bayes classifier and also suggested potentials avenues to improve their model. \textcolor{black}{Ozbay  and Alatas \cite{ozbay2020fake} proposed a combination of text mining techniques and supervised artificial intelligence algorithms for the task of fake news detection. They showed that the best mean values in terms of accuracy, precision, and F-measure are obtained from the Decision Tree algorithm. Apart from textual features, visual features have also been employed for fake news detection. \cite{zhou2020} proposed a similarity-aware fake news detection method which utilizes the multi-modal data for effective fake news detection. On the similar lines, Varshney  and Vishwakarma \cite{varshneyunified} developed a click-bait video detector which is another prevalent form of online false content. Despite the success of supervised models, news spreads on social media at very high speed when an event happens, only very limited labeled data is available in practice for fake news detection}. Some studies such as \cite{helmstetter2018weakly, gravanis2019behind} have been involved around weakly supervised learning for fake news detection. In similar directions, Yu et al. \cite{yu2017constrained} used constrained semi-supervised learning for social media spammer detection, while Guacho et al. \cite{guacho2018semi} used tensor embeddings to form a semi-supervised model for content based fake news detection. \textcolor{black}{Dong  et al. \cite{dong2020two}  proposed a two-path deep semi supervised learning for timely detection of fake news. They verified their system on two datasets and demonstrated effective fake news detection. Vishwakarma et al. \cite{vishwakarma2019detection}  analysed the credible web sources and proposed a reality parameter for effective fake news prediction. Varshney et al. \cite{varshney2020hoax}  developed an automated system Hoax-News Inspector for real time prediction of fake news. They used content resemblance over web search results for authenticating the credibility of news articles. Recently, Patwa et al. \cite{patwa2021overview} prepared an English COVID-19 fake news dataset \cite{patwa2020fighting} and a Hindi hostile post dataset \cite{bhardwaj2020hostility}. 
A few recent studies \cite{meel2019fake, bondielli2019survey, zhou2020survey} have provided extensive literary surveys by investigating datasets, features and models along with potential future research prospects for fake news detection.}

\textbf{Semi-supervised models for text classification:}
Semi-supervised learning (SSL) is proved to be powerful for leveraging unlabelled data when we lack the resources to create large-scale labelled dataset. Prior research on semi-supervised learning can broadly be divided into three classes-- multi-view, data augmentation and transfer learning \cite{sachan2019revisiting}. The objective of multi-view approaches is to use multiple views of labelled as well as unlabelled data. Johnson and Zhang \cite{johnson2015semi} obtained multiple views for text categorisation by learning embedding of small text regions from unlabelled data and integrating them to a supervised model. Gururangan et al. \cite{gururangan2019variational} and Chen et al. \cite{chen2019variational} leveraged variational autoencoders in the form of sequence-to-sequence modelling on text classification and sequential labelling. Data augmentation approaches involve augmenting either the features or labels. Nigam et al. \cite{nigam2000text} classified the text using a combination of Naive Bayes and Expectation Maximisation algorithms and demonstrated substantial performance improvements. Miyato et al. \cite{miyato2016adversarial} utilized adversarial and virtual adversarial training to the text domain by applying perturbations to the word embeddings. Chen et al. \cite{chen2020mixtext} introduced MixText that combines labelled, unlabelled and augmented data for the task of text classification. They interpolated text in hidden space using Mixup \cite{zhang2017mixup} to create a large number of augmented training samples. Xie et al. \cite{xie2019unsupervised} used advanced augmentation methods (RandAugment and back-translation) to effectively noise unlabelled examples. Transfer learning approaches aim to initialise task-specific model weights with the help of pre-trained weights on auxiliary tasks. Dai and Le \cite{dai2015semi} used a sequence autoencoder, which reads the input sequence into a vector and predicts the input sequence again to use unlabelled data for improving sequence learning with recurrent networks. \textcolor{black}{Hussain and Cambria \cite{hussain2018semi} employed a semi-supervised model based on the combined use of random projection scaling, and support vector machines to perform reasoning on a knowledge base. They showed a significant improvement in emotion recognition and polarity detection tasks over the state-of-the-art methods.} Howard et al. \cite{howard2018universal} proposed the Universal Language Model Fine-tuning (ULMFiT), which has been proved as an effective transfer learning method for various NLP tasks. Both studies \cite{howard2018universal,dai2015semi} showed the improvement in the performance of text classification using transfer learning.

\textcolor{black}{
The most of the aforementioned methods for fake news detection are tested on datasets with high volume of labelled data. Moreover, when multiple features are considered, their optimal combination is not explored. There is no published work related to COVID-19 fake news detection. We strive to address these issues by first introducing the novel \CTF\ dataset and then leveraging the unlabelled data in order to reduce the vast dependency on the labelled data in our proposed Cross-SEAN model. We also employ cross-stitch for optimal combination of inputs into various sections of the model and show interesting analysis.}

\section{Dataset Collection and Annotation \label{sec:dataset}}

\begin{table}[]
\centering
\scalebox{0.8}{
\begin{tabular}{|c|c|c|c|c|c|c|c|c|}
\hline
\multicolumn{9}{|c|}{Dataset Collection and Labelling} \\ \hline
\multicolumn{3}{|c|}{\begin{tabular}[c]{@{}c@{}}Using Hashtags\\ and Keywords\end{tabular}} & \multicolumn{3}{c|}{\begin{tabular}[c]{@{}c@{}}Using Statements and\\ Tweets from Organisations\end{tabular}} & \multicolumn{3}{c|}{Using URLs} \\ \hline
\multirow{2}{*}{\begin{tabular}[c]{@{}c@{}}Major\\ Keywords\end{tabular}} & \multicolumn{2}{c|}{No. of Tweets} & \multirow{2}{*}{\begin{tabular}[c]{@{}c@{}}Major\\ Sources\end{tabular}} & \multicolumn{2}{c|}{No. of Tweets} & \multirow{2}{*}{\begin{tabular}[c]{@{}c@{}}Major\\ Services\end{tabular}} & \multicolumn{2}{c|}{No. of Tweets} \\ \cline{2-3} \cline{5-6} \cline{8-9} 
 & Fake & Genuine &  & Fake & Genuine &  & Fake & Genuine \\ \hline
bioweapon & 4978 & 0 & WHO & 3395 & 4700 & Snopes & 1696 & 1650 \\ \hline
vaccine & 3620 & 221 & CDC & 1649 & 2195 & PolitiFact & 1484 & 2250 \\ \hline
trump & 2874 & 439 & NIH & 2231 & 1705 & FactCheck & 1060 & 1500 \\ \hline
china & 2677 & 515 & CPHO & 582 & 470 & TruthOrFiction & 1042 & 1895 \\ \hline
WHO & 493 & 4018 & PHE & 391 & 425 & - & - & - \\ \hline
at home & 0 & 4552 & HHS & 405 & 2255 & - & - & - \\ \hline
\end{tabular}}
\caption{ \textcolor{black}{Different attributes including keywords, hashtags, and sources of statements and URLs along with the respective number of tweets they are responsible for. The table compiles the numeric details of Section \ref{sec:dataset}. Here, WHO: World Health Organisation, CDC: Centers for Disease Control, NIH: National Institute of Health, CPHO: Central Public Health Office, PHE: Public Health England, HHS: Human and Health Services.}}\label{tab:data_statistics}
\end{table}

In this section, we introduce our novel dataset, called \textbf{\CTF}\ ({\bf C}OVID-19 {\bf T}witter {\bf F}ake News). The formation of this dataset underwent four stages mentioned below.

{\bf Stage 1. Segregating COVID-19 related tweets:} 
Multiple COVID-19 Twitter datasets (unlabelled) have recently been made public on Kaggle and other sources; among them, we used the datasets released by \cite{kaggleCarl}, \cite{kaggleShane}, and \cite{kaggleSVEN}. Alongside, there exist a few publicly available datasets containing COVID-19 related tweet IDs being released everyday in chronological order. We collected the tweet IDs from \cite{gitEchen} and \cite{QCRI}. Due to the hydrating process (which is time consuming) and the non-existence of fake tweets (as Twitter deletes them upon identification), the tweet IDs did not turn out to be very useful. However, we still considered them in our dataset to learn the language semantics explained in the subsequent sections. We also collected tweets using the Twitter API based on some predefined hashtags (e.g., `WHO', `covid19', `wuhan', `bioweapon', etc.). Since the genuineness of news correlates to the credibility of the source, we collected tweets published by the aforementioned governmental health organisations and gathered their official Twitter IDs. We extracted tweets from these accounts and considered them genuine.

\textbf{Stage 2. Collecting COVID-19 supporting statements:}
There exist fact checking sites which analyse popular news across social media and label them as fake or genuine based on verified sources. We crawled various fact-checking sites such as \textit{Snopes}, \textit{PolitiFact}, \textit{FactCheck} and \textit{TruthOrFiction} for content related to COVID-19. We extracted URLs, the content of URLs and their corresponding labels (genuine or fake) from the fact checking websites. To support this data, more genuine URLs were extracted from the Twitter accounts of the official health bodies. To increase public awareness about any widely accepted misinformation, governmental bodies across the world have setup specific web pages\footnote{\url{https://www.who.int/emergencies/diseases/novel-coronavirus-2019}} \footnote{ \url{https://www.cdc.gov/coronavirus/2019-ncov/index.html}} \footnote{\url{https://www.coronavirus.gov/}}, which are also scraped. This stage resulted in a bulk amount of data related to the content and URLs which are known to be fake/genuine and act as the supporting statements for the next stage.

\textbf{Stage 3: Filtering genuine and fake tweets:}
We assumed that when a fake or genuine URL is being shared, all the tweets accompanying the URL also belong to the same class as URLs are generally added in support to the text. Based on this, a total of $5.3K$ and $7.5K$ tweets were labelled as fake and genuine, respectively. Although this assumption may garner some unwanted noise since a tweet might contradict the opinion presented to the referred URL, on manual inspection we found out that this assumption surprisingly held true for most of the cases, as elaborated in the next section. In addition, all the tweets posted by governmental health organisations related to COVID-19 with specific hashtags as mentioned above, form a majority of our genuine data. This is based on the assumption that such health organisations post content which either curb fake news or are genuine in itself. We gathered $10K$ genuine tweets via this method. Next, we used the pre-processed tweet texts with two Transformer models, BERT \cite{Reimers_2019} and RoBERTa \cite{liu2019roberta}, to populate the dataset further. BERT is used to generate embeddings of both tweet text and the supporting statements collected and cosine distance is computed with a high threshold of $0.9$ to label the tweet into genuine or fake based on the polarity. This step resulted in $9.7K$ tweets labelled as fake. For RoBERTa, we used the fine-tuned version on the Stanford Natural Language Inference (SNLI) Corpus \cite{snli:emnlp2015}; this allowed to take in a pair of sentences and check if they are contradicting, neutral or entailing. We formed pairs of tweets and supporting statements to identify genuine or fake tweets based on contradicting and entailing results. This approach gave us an extensive set of $10.6K$ fake tweets.

{\bf Stage 4: Human annotation:} We performed manual verification of a part of $45,261$ labelled tweets ($26,706$ fake, $18,555$ genuine) obtained from Stage 3. We employed three human annotators, who are experts in social media and have significant expertise in fact verification, to verify the labels. The annotators ended up annotating $16,000$ tweets ($8000$ fake and $8000$ genuine) with an inter-annotator's agreement of $0.82$ (Krippendorf's $\alpha$) with the following instructions provided: \label{human}

\begin{itemize}
    \item A tweet is  considered to be `fake' if and only if:
    \begin{itemize}
        \item It contradicts or undermines facts from a pre-defined list. Note that a combined list was made from the aforementioned genuine sources.
        \item It supports or elevates a commonly identified misinformation.
        \item It is written in the form of sarcasm or humour, but promotes a misleading statement. 
    \end{itemize}
    \item Other tweets which do not satisfy any of the above, would be either unlabelled or genuine, as per the annotator's discretion.
    \item If the tweet text in itself does not provide enough context to annotate with confidence, the annotators could refer to the tweet and user features.
\end{itemize}

On further observation, it is found that an average of $92\%$ labels given by the automated techniques from Stage-3 matched the labels given by the human annotators for $16,000$ samples. Thus, despite using a fully-automated and fast annotation pipeline, which allowed us to have a relatively large labelled corpus, only a noise of $8\%$ exists.

During cross-validation, we use 20\% of the human-verified tweets for testing, and remaining 80\% tweets along with the unverified tweets\footnote{It may plant some noise in the training set which a sophisticated classifier should ignore while being trained.} constitute the training set. We maintain the same distribution of fake and genuine tweets present in the entire dataset in both the training and test sets.

\begin{figure}[htp]
  \centering
  \subfigure[Hashtags]{\includegraphics[height=3cm, width=0.3\textwidth, clip = true]{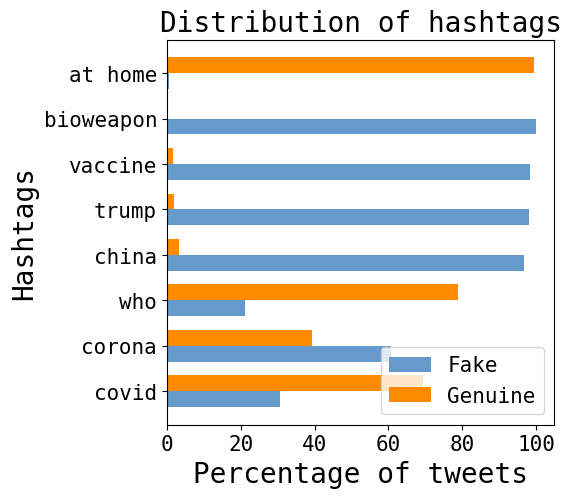}\label{fig:ana_hs}}\quad
  \subfigure[Sentiment]{\includegraphics[height=3cm, width=0.3\textwidth, clip = true]{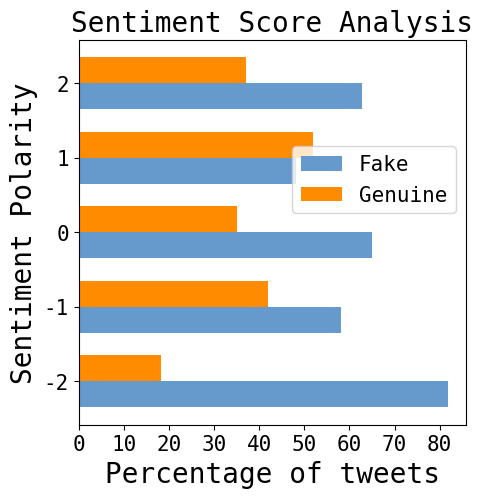}\label{fig:ana_ss}}\quad
  \subfigure[Likes]{\includegraphics[height=3cm, width=0.3\textwidth, clip = true]{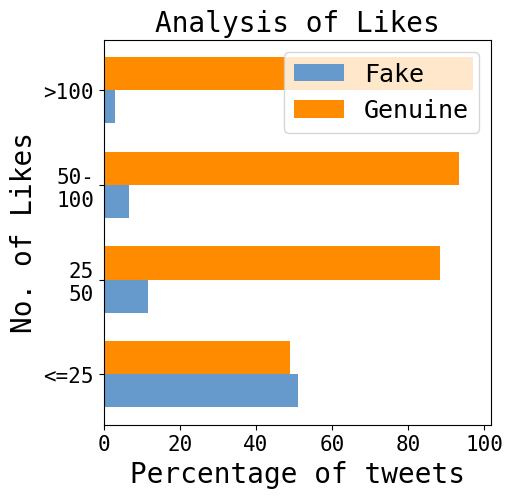}\label{fig:ana_likes}}\quad
  \caption{(a), (b) and (c) show the distribution of hashtags, sentiment and likes across the tweets, respectively.}
  \label{fig:features_l_ul}
\end{figure}

\section{Dataset Analysis \label{sec:analy}}

\textcolor{black}{In Table \ref{tab:data_statistics}, we show major keywords, statements and tweets from organisations and URLs used and the number of tweets that are labelled as fake and genuine.}

\textbf{Presence of hashtags:} Hashtags have long been an important tool on Twitter to organise, sort, follow and spread tweets. Our dataset consists of a total of $955$ and $2,231$ unique hashtags in genuine and fake tweets, respectively. We tabulate the distribution of hashtags for tweets in Fig. \ref{fig:ana_hs}. It is evident that `\#WHO' is more prominent in genuine tweets. The vast number of tweets containing `\#china' and `\#bioweapon' are fake tweets. Interestingly, the appearance of `\#trump' hashtag in the fake tweets is much higher than the genuine tweets, pointing towards the tendency of politicisation amongst fake tweets. Even though the vaccine for COVID-19 is still under development, the recurrent use of `\#vaccine' in fake tweets may suggest the tendency of spreading rumours with false remedies. The dominance of hashtags such as `\#togetherathome', `\#stayhome' and `\#socialdistancing' in the genuine tweets suggests that they might have been used to spread positive social messages.


\textbf{Presence of URLs:} To account for prevalence of misinformation, we analyse the URLs present in our entire dataset. A total of $14,830$ genuine and $8,761$ fake tweets contain at least one URL, thus averaging to $0.87$ and $0.35$ URLs per genuine and fake tweet, respectively. The contrast between the numbers may suggest that in general, genuine tweets have a higher tendency of supporting the claims.

\textbf{Presence of multimedia:} Twitter supports three types of media formats in a tweet-- photo (P), video (V) and GIF (G). However, it supports only one type of media in a particular tweet with a limit of four photos and only one video/GIF. In our dataset, fake tweets contain a total of $2,491$ media files (2036P, 381V, 74G) across $2,344$ tweets, with an average of $0.0988$ per tweet, while genuine tweets contain $1,473$ media (1129P, 339V, 5G) with an average of $0.0834$.

\begin{figure}[!t]
  \centering
  \subfigure[TT-L]{\includegraphics[trim=1cm 1cm 0cm 1cm, width=0.3\linewidth, clip = true]{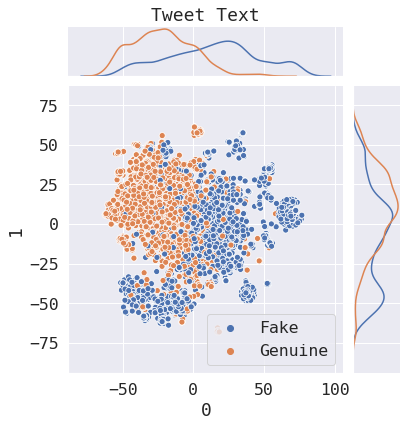}\label{fig:ana_tsne_labelled_tweet_text}}\quad
  \subfigure[TF-L]{\includegraphics[trim=1cm 1cm 0cm 1cm, clip = true, width=0.3\linewidth]{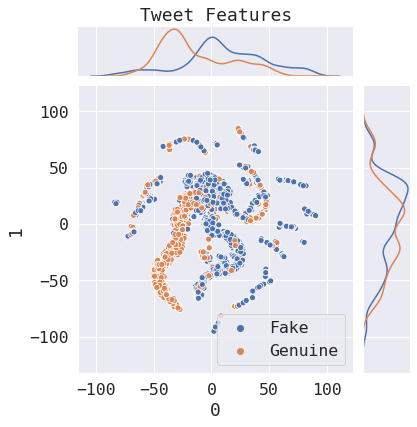}\label{fig:tf_l}}\quad
  \subfigure[UF-L]{\includegraphics[trim=1cm 1cm 0cm 1cm, clip = true, width=0.3\linewidth]{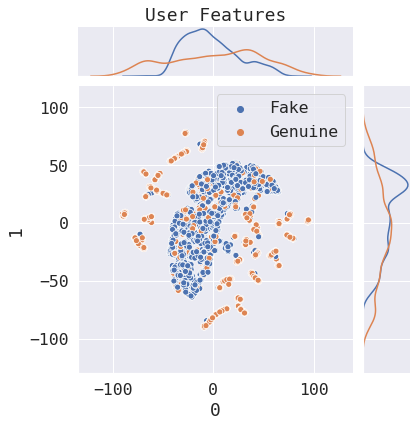}\label{fig:uf_l}}\quad
  \caption{(a), (b) and (c) show the t-SNE visual representations of tweet text, tweet features and user features of the labelled data, respectively. Here, TF $\rightarrow$ Tweet Features, UF $\rightarrow$ User Features and L $\rightarrow$ Labelled Data.}
  \label{fig:features_l_l}
  \vspace{-5mm}
\end{figure}

\textbf{Sentiment of tweets:}
To obtain overall sense of public opinion related to COVID-19, we analyse sentiment of the tweets \cite{CHAKRABORTY2020106754} using the texblob\footnote{\url{https://textblob.readthedocs.io/en/dev/}} tool. Fig. \ref{fig:ana_ss} shows that in the highly negative (-2) and neutral (0) sentiment zones, fake news are grouped more than the genuine news. The average sentiment polarity for fake tweets is $0.05$ compared to $0.096$ in genuine tweets, on a scale of -2 to 2, as shown in Fig. \ref{fig:ana_ss}. \textcolor{black}{Similar results were also obtained from latest state-of-the-art polarity classification methods for long reviews and short tweets \cite{cambria2017affective, akhtar2020intense, basiri2021abcdm}.}

\textbf{Likes and retweets:} 
The existing propagation based approaches \cite{granik2017fake, jang2019fake} showed the significance of likes and retweets for fake news detection. The average number of likes per genuine tweet is found to be $142.65$, which is significantly higher than that ($4.25$) of fake tweet. The tweet-wise data of likes is summarised in Fig. \ref{fig:ana_likes}. The large number tweets of popular public health organisation explains the higher average likes per genuine tweet. About $64\%$ of fake tweets in our dataset are retweets of some other tweet, $8\%$ of the fake retweets are quoted with the comments, and $35\%$ of genuine tweets are retweets with 8\% of them being retweets with comment.

\begin{figure}[!t]
  \centering
  \subfigure[TT-UL]{\includegraphics[width=0.3\linewidth, clip = true]{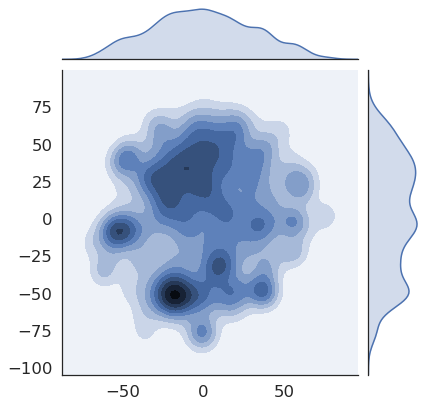}\label{fig:ana_tsne_unlabelled_tweet_text}}\quad
  \subfigure[TF-UL]{\includegraphics[trim=1cm 1cm 0cm 0cm, clip = true, width=0.3\linewidth]{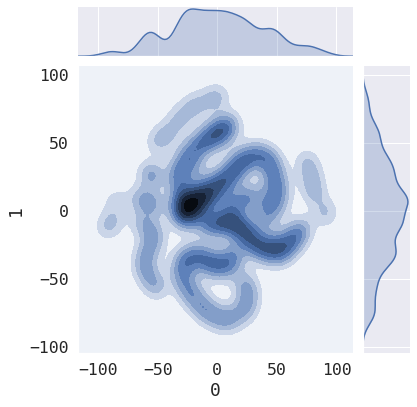}\label{fig:tf_ul}}\quad
  \subfigure[UF-UL]{\includegraphics[trim=1cm 1cm 0cm 0cm, clip = true, width=0.3\linewidth]{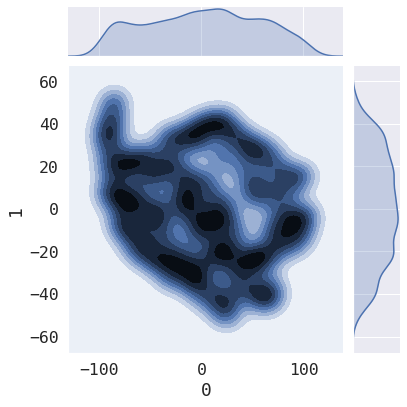}\label{fig:uf_ul}}\quad
  \caption{(a), (b) and (c) show the t-SNE visual representations of tweet text, tweet features and user features of the unlabelled data, respectively. Here, TF $\rightarrow$ Tweet Features, UF $\rightarrow$ User Features and UL $\rightarrow$ Unlabelled Data.}
  \label{fig:features_ul_ul}
  \vspace{-5mm}
\end{figure}

\textbf{Visual representations:} 
We show t-SNE visual representations of labelled and unlabelled tweets on tweet text, tweet features and user features in Figs. \ref{fig:features_l_ul}, \ref{fig:features_l_l} and \ref{fig:features_ul_ul}. Fig \ref{fig:features_l_ul} shows tweet text representations on labelled and unlabelled data. Sentence BERT is used to convert the tweet text to vector form. While the overlap of genuine and fake tweets can be observed from Fig. \ref{fig:ana_tsne_labelled_tweet_text}, the polarisation of topics can be observed from the unlabelled data from Fig. \ref{fig:ana_tsne_unlabelled_tweet_text}. 
Certain user features and tweet features are identified and are mentioned in \Cref{feature_names}; these are in turn used for the visualisations on labelled and unlabelled data in Figs. \ref{fig:features_l_l} and \ref{fig:features_ul_ul} respectively. The polarisation in Fig. \ref{fig:features_ul_ul} supports the same in Fig. \ref{fig:ana_tsne_unlabelled_tweet_text}. The labelled representation shows high non-linear overlap and indicates the complexity of the classification task. 

\section{Cross-SEAN: Our Proposed Method \label{sec:model}}
\begin{figure}
    \centering
    \includegraphics[scale=0.3]{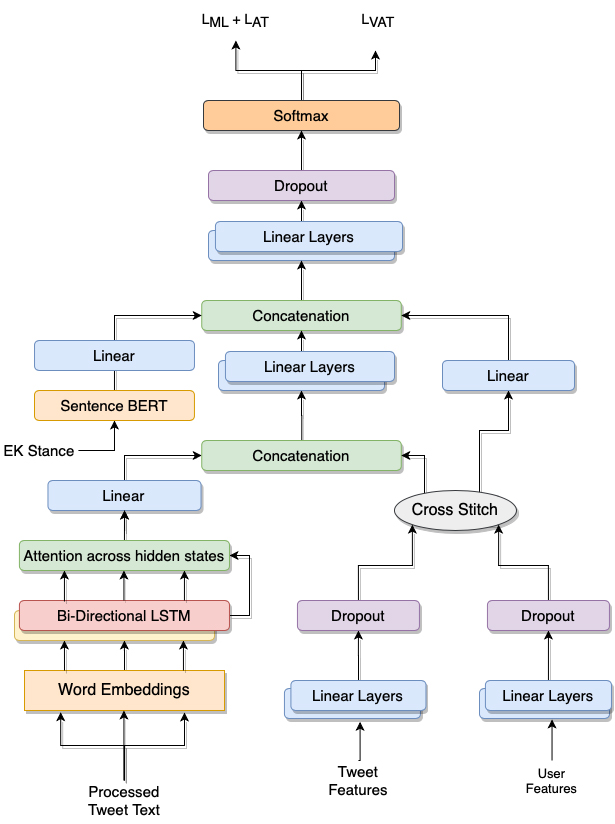}
    \caption{A schematic diagram of Cross-SEAN.}
    \label{fig:cross-sean}
\end{figure}

In this section, we describe Cross-SEAN\footnote{{\bf Cross}-Stitch based {\bf S}emi-Supervised {\bf E}nd-to-End {\bf A}ttention {\bf N}eural Network} for fake news detection. We explain individual components of the model, followed by the training strategy. Fig. \ref{fig:cross-sean} shows the architecture of Cross-SEAN.

\subsection{Explicit Tweet and User Features and External Knowledge} \label{feature_names}
Monti et al. \cite{monti2019fake} showed that content, social context or propagation in isolation is insufficient for neural models to detect fake news. Hence, we employ additional features related to both the users and tweets along with the content of the tweets. For the tweet features (TFs), we consider the attributes available in the tweet object and some handcrafted features from the tweet, amounting to a total of 10 features -- \textit{number of hashtags, number of favourites, number of retweets, retweet status, number of URLs present, average domain score of the URL}({\it s}), \textit{number of user mentions, media count in the tweet, sentiment of the tweet text, counts of various part-of-speech tags} and \textit{counts of various linguistic sub-entities}.

Polarisation of users on similar beliefs is widely observed on Twitter \cite{yardi2010dynamic}. To capture this, we extract 8 features for each corresponding user (UFs) -- \textit{verified status, follower count, favourites count, number of tweets, recent tweets per week, length of description, presence of URLs} and \textit{average duration between successive tweets}.

These features can provide additional information of the user characteristics and their activities. These not only help the model identify bots and malicious fake accounts, but also help recognise a pattern amongst users who post false and unverified information. 

On visualising the tweet and user features on labelled and unlabelled data in Fig. \ref{fig:features_l_ul}, we observe the formation of clusters of similar tweets, indicating the polarity of the tweets. From Fig. \ref{fig:uf_l}, we also observe that users posting fake tweets tend to form a cluster, and users posting genuine tweets are scrambled across the whole feature space. Few features of user posting genuine tweets are highly similar to the features of users posting fake tweets, thus overlapping with the fake tweet cluster. From Fig. \ref{fig:features_ul_ul}, the unsupervised user features show the dense polarity across the whole latent space, while the tweet features are wide spread, showing the diverse set of attributes in our unlabelled data. These features are further used in the classification.

Feature based neural models learn a generalised function from a limited manifold of the training data, and thus have a tendency to perform poorly when the topics are variant. To overcome this in Cross-SEAN, we use external knowledge, for the content relating to tweet text, as an input to the model. We use classical text processing techniques to find a shortened contextual form of the tweet text and use it as a query to retrieve the top Google Search results, sorted in accordance to relevance\textcolor{black}{(\cite{vishwakarma2019detection, varshney2020hoax, malhotra2020classification} )}. From each web-article returned from the search, a particular number of text sentences are retrieved which are the closest to the original tweet text, as measured using cosine similarity of the BERT Sentence Embeddings \cite{Reimers_2019} of the two. This is done until $k$ (=10, by default) sentences are retrieved for the tweet.

In addition to this, we make use of the large amount of unlabelled data ($21.85$M) available in \CTF--

\begin{itemize}
    \item We use one-half of the unlabelled data to fine-tune word embeddings to encode the tweet text. We expect this to help the model learn the linguistic, syntactic and contextual composition of not only general Twitter Data but also the domain data, i.e., the COVID-19 pandemic in case of \CTF.
    \item We leverage the other half of the unlabelled data for unsupervised training using an additional adversarial loss. Experimental results presented in Section \ref{sec:ablation} show that doing this reduces stochasticity and makes the model more robust with the nature of adversarial training.
\end{itemize}

We elaborate on  various components of the model architecture and the training intricacies in the following sections.

\subsection{Model Architecture}

Our entire training data is composed of labelled and unlabelled samples, denoted by $X_L$ and $X_U$ respectively. $X_L$ consists of a total of $n_L$ data points: ($x_L^1$, $y_L^1$), ($x_L^2$, $y_L^2$), $\cdots$, ($x_L^{n_L}$, $y_L^{n_L}$), where $x_L^i$ is the $i^{th}$ tweet and $y_L^i$ is its  label. $X_U$ consists of a total of $n_U$ unlabelled data points: $x_U^1$, $x_U^2$, ..., $x_U^{n_U}$. In both the cases, each input sample, $x_K^i$ (for $K \in (L, U$)) comprises four input sub-sets -- tweet text ($x_{TT}^i$), external knowledge text ($x_{EK}^i$), tweet features ($x_{TF}^i$) and user features ($x_{UF}^i$).\\
In each pass through our model, these four inputs are encoded separately as described below.

\textbf{Encoding textual data:}
The tweet text of sequence length $N$ is represented as a one-hot vector of vocabulary size $V$. A word embedding layer $E \in R^{V \times D}$ transforms the one-hot vector into a dense tensor $e$ $\in R^{N \times D}$ consisting of ($e^1, e^2, ..., e^N$). These token vectors are further encoded using a Bidirectional LSTM, the forward and backward layers of which process the $N$ vectors in opposite directions.

The forward LSTM emits a hidden state $h_{ft}$ at each time-step, which is concatenated with the corresponding hidden state $h_{bt}$ of the backward LSTM to produce a vector $h_t \in R^{(2\times H)}$, 

\textcolor{black}{\begin{equation}
    h_t = h_{ft} \oplus h_{bt}, \forall t \in [1, N]
\end{equation}}

where $H$ is the hidden size of each LSTM layer. 

At each layer, a final state output $f_{k} \in R^H$ is also obtained ($\forall$ $k \in (f, b)$). 

At this stage, a net hidden vector $h$ containing $N$ hidden vectors from the two LSTM layers is combined with the final state vector $f$ using attention across the hidden states, given as:
\textcolor{black}{\begin{equation}
v = \sum_{j=1}^{N} \alpha_{ij} h_{j}; \alpha_{ij} = Softmax(h_i \bullet f_j)
\end{equation}}

where, 
\textcolor{black}{\begin{equation}
    f = f_{f} \oplus f_{b}, f \in R^{2 \times H}
\end{equation}}

\textcolor{black}{\begin{equation}
    h = h_1 \cdots \oplus \cdots h_N, h \in R^{N, (2\times H)}
\end{equation}}

We refer vector $v$ obtained after attention across the hidden states as $v_{TT}$, representing the encoded feature of the tweet text. 

In addition to this, we use Sentence BERT \cite{Reimers_2019} to find contextual embedding $e_{EK}$ of the external knowledge corresponding to each input batch. We do this considering the vast difference between our tweet text input and the external knowledge text. The $e_{EK}$ vector is then passed through a linear layer to obtain an encoded representation $v_{EK}$ of the external knowledge.

\textbf{Encoding tweet and user features:} 
As shown in Fig. \ref{fig:features_l_ul}, we follow a highly concurrent yet distinct mechanism to encode both tweet and user features. Firstly, $x_{TF} \in R^{K_t}$ and $x_{UF} \in R^{K_u}$ are passed through separate linear layers which interpolate them to higher dimensional dense feature vectors $v_{TF} \in R^{K_T}$ and $v_{UF} \in R^{K_U}$, respectively.  
As both $x_{TF}$ and $x_{UF}$ are handcrafted, we employ cross-stitch units, which not only allow the model to learn the best combination of inputs from both the features and share across multiple layers in the network, but also introduce a common gradient flow path through the non-linear transformation. The transformation produced by cross-stitch is as follows:

\begin{equation}
    v_{j}' = \alpha_{ij} \bullet v_{j} + \beta_{i}, \forall i, j \in (1, K_T + K_U)
    \label{eq:cross}
\end{equation}

where $\alpha_{ij}$ and $\beta_{i}$ denote the weights of the fully connected layer performing the cross-stitch operations. 

The two outputs of the cross-stitch are denoted by $v_{TU}$ and $v_{UT}$, respectively\footnote{The first letter in the subscript of $v$ denotes feature vector assuming that it contains most information from the same vector.}. Note that the shape of the two vectors remains unchanged after this transformation. 

\begin{figure}
    \centering
    \includegraphics[width=0.5\linewidth]{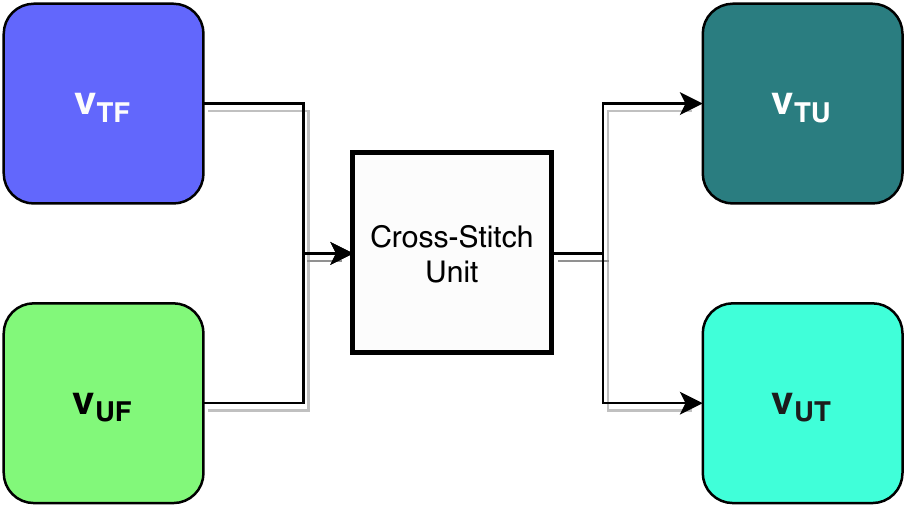}
    \caption{Working of a cross-stitch unit. Here, the notation is as  defined in Eq. \ref{eq:cross}. Note that the weights of the linear layers in the Cross-stitch unit are initialised with a unit matrix.}
    \label{fig:cross}
    \vspace{-5mm}
\end{figure}

\textbf{Connected components in Network:} 
We concatenate $v_{TT}$ and $v_{TU}$, which are the transformed feature vectors of the tweet text $x_{TT}$ and tweet features $x_{TF}$, respectively. This produces $v_{T} = v_{TT} \oplus v_{TU}$, a concatenated representation of all textual features. This is done considering the inherent similarity between the tweet text and the tweet features over user features. We then perform affine transformations of the three vectors, $v_{T}$, $v_{EK}$ and $v_{UT}$, through separate feed-forward linear layers and concatenate to obtain the final decoded vector $v$, effectively containing transformed feature representations from all the inputs. The vector $v$ is then down-scaled using a fully-connected network, regularized using dropout before finally obtaining the probability distribution across the two classes.

\textcolor{black}{\begin{equation}
    p(y | x; \theta) = Softmax(v') = Softmax(||v_{T}'||v_{UT}'||')
\end{equation}}

where, $v'$ represents the transformed vector after it passes through the respective feed forward sub-network, and $\theta$ represents the model parameters at the current time (from now on, we refer to this as $f(x)$).

\subsection{Training Strategies}
For training our model, we use a mixed objective function, which is a weighted sum of both supervised and unsupervised losses:
\textcolor{black}{\begin{equation}
    L_{mix} = \lambda_{ML}L_{ML} + \lambda_{AT}L_{AT} + \lambda_{VAT}L_{VAT}
\end{equation}}

The losses are as follows: (i) $L_{ML}$ represents maximum likelihood loss and minimizes the loss between the predicted and true labels. (ii) Additionally, we use the Adversarial Training Loss $L_{AT}$, which introduces a regularization with model training by adding a denoising objective \cite{miyato2016adversarial}. The goal through this training is to make the model robust to adversarial perturbations in the input. We find this specially useful for fake news detection as it allows the model to attend to a wide spectrum of tweets with minor variations to improve the generality. An adversarial signal $r_{adv}$, defined in terms of the $L_2$ norm on the gradient $g_L$, with current model parameters is used to perturb the word embedding inputs $e$ of $x_{TT}$, $e\star = e + r_{adv} $, even when this perturbation depends upon the gradient computed over the output w.r.t all the labelled inputs $x_L$. The $L_{AT}$ objective function in Eq. \ref{eq:at} is given as a modification of $L_{ML}$ (Eq. \ref{eq:ml}). (iii) It can be observed that the above two objectives require us to know the true label of the data input, thus pertaining to the labelled data only. Here, to expand the concept of adversarial training to unlabelled data, we make use of virtual adversarial training loss $L_{VAT}$, which too is aimed to add robustness against adversarial inputs. Just as in Eq. \ref{eq:at}, we apply the perturbation on the word embedding $e$, except $r_{adv}$ is now defined as in Eq. \ref{eq:rvadv}. $\delta$ represents a small random perturbation vector \cite{miyato2016adversarial}, using a 2nd-order Taylor series expansion followed by the power iteration method. The VAT loss is then defined as in Eq. \ref{eq:vat}. We denote $f(x) = h(E(x))$, where $E(x) \in R^{N \times D}$ is the word embedding vector.

\vspace*{-0.2cm}
\begin{equation}\small
    L_{ML} = \frac{-1}{n_{L}}\sum_{i=1}^{n_{L}}y_i \log(f(x_i)) + (1-y_i) \log(1-f(x_i))
    \label{eq:ml}
\end{equation}
\vspace*{-0.15cm}
\begin{equation}\small
    r_{adv} = -\epsilon _L/||g_L||_2; g_L = -\nabla_{x_{L}} \log(f(x_L))
    \label{eq:radv}
\end{equation}
\vspace*{-0.2cm}
\begin{equation}\small
    L_{AT} = \frac{-1}{n_{L}}\sum_{i=1}^{n_{L}} P + Q
    \label{eq:at}
\end{equation}
where, 
\textcolor{black}{\begin{equation}
    P = y_i \log(h(E(x_i)) + r_{adv})
\end{equation}}
\textcolor{black}{\begin{equation}
    Q = (1-y_i) \log(1-f(h(E(x_i) + r_{adv})))
\end{equation}}
\vspace*{-0.2cm}
\begin{equation}\small
    r_{v-adv} = \epsilon g/||g||_2; g = -\nabla_{x} KL[f(x) || h(E(x) + \delta)]
    \label{eq:rvadv}
\end{equation}
\vspace*{-0.23cm}
\begin{equation}\small
    L_{VAT} = \frac{1}{n_{L} + n_{U}}\sum_{i=1}^{n_{L} + n_{U}} KL[f(x) || h(E(x) + r_{v-adv})]
    \label{eq:vat}
\end{equation}

\begin{table*}[ht]
\centering
\scalebox{0.8}{
    \begin{tabular}[H]{ |l||c|c|c|c||c|c|c|c|}
     \hline
     Model&\multicolumn{4}{c||}{Features used by the model}&\multicolumn{4}{c|}{Performance}  \\
     \cline{2-5}
     \cline{6-9}
     &TT &TF & UF & UL & Accuracy  &Precision & Recall &F1 Score\\
     \hline
     MTL &\checkmark &\checkmark & & &0.79 &0.77 &0.82 &0.79 \\
     \hline
     1HAN &\checkmark & & & &0.89 &0.60 &0.87 &0.71\\
     \hline
     16HLT-HAN &\checkmark & & & &0.87 &0.68 &0.86 &0.76 \\
     \hline
     3HAN &\checkmark & & &\checkmark &0.89 &0.77 & 0.82 &0.80 \\
     \hline
     CSI &\checkmark & &\checkmark & &0.87 &0.80 &0.91 &0.85 \\
     \hline
     dEFEND &\checkmark & &\checkmark & &0.89 &0.83 &0.89 &0.86 \\
     \hline
     MixText &\checkmark & & &\checkmark &0.87 &0.83 &0.84 &0.84 \\
     \hline
     \textbf{Cross-SEAN} &\checkmark &\checkmark &\checkmark &\checkmark &\textbf{0.954} &\textbf{0.946} &\textbf{0.961} &\textbf{0.953}\\
     \hline
    \end{tabular}}
\caption{Features used by the competing models and performance comparison on \CTF\ (TT: Tweet Text, TF: Tweet Features, UF: User Features, UL: Unlabelled Data).}
\label{tab:ress1}
\vspace{-5mm}
\end{table*}

\section{Experimental Setup and Results \label{sec:experiment}} 
All our experiments were performed on a single 16 GB Nvidia Tesla V-100 GPU.
Our base model is a single layer Bi-LSTM with a maximum sequence length of 128 and a hidden dimension of 512. We performed experiments with a wide range of embedding sizes ranging from 128 to 768 and found the best results with 300 dimensions. We initially fine-tuned the word embeddings on $\sim 10M$ unlabelled tweet texts before using them for training.
We used the $Adam$ optimiser for all our experiments with a learning rate of 0.001, $\beta_1=0.90$, $\beta_2=0.98$ and a decay factor of 0.5. We used dropout with $p_{drop}$ of 0.3 in all our feed-forward networks, where the number of layers exceeds 2. Early stopping with a patience of $20$ was also used along with gradient clipping with a maximum $L_2$ norm of 1. 
We kept $\lambda_{ML}$, $\lambda_{AT}$ and $\lambda_{VAT}$ as 1.

\subsection{Comparative Evaluation}
We compare Cross-SEAN with seven state-of-the-art methods described as follows.
{\bf MTL} \cite{arora2019multitask} uses a multitask learning framework by leveraging soft parameter sharing on classification (primary) and regression (secondary) tasks based on tweet text and tweet features.
1HAN and 3HAN \cite{singhania20173han} use hierarchical attention based GRU networks. {\bf 1HAN} is the base version of 3HAN, where {\bf 3HAN} uses 3-level hierarchical attention for words, sentences and headlines learning in a bottom up manner. {\bf 16HLT-HAN} \cite{yang2016hierarchical} uses hierarchical structure by applying attention mechanism at both word and sentence levels.
{\bf CSI} \cite{ruchansky2017csi} uses a three module approach that consists of \textit{Capture}, \textit{Score} and \textit{Integrate}, combining what they define as the three common characteristics among fake news, i.e., \textit{text}, \textit{response} and \textit{source} to identify misinformation. Furthermore, we also use {\bf dEFEND} \cite{cui2019defend} as a baseline, which uses a GRU-based word-level and sentence-level encoding along with a module for sentence-comment co-attention. {\bf MixText} \cite{chen2020mixtext} is a semi-supervised approach that produces results by leveraging large amount of training samples and interpolating text in hidden space. 

Table \ref{tab:ress1} shows that Cross-SEAN outperforms all the baselines by a margin of at least more than 6\% accuracy and 9\% F1 Score, with dEFEND being the best baseline.

\begin{table}
\centering
    \scalebox{0.8}{
    \begin{tabular}[H]{|l|c|c||c|c|}
     \hline
     \multicolumn{3}{|c||}{Objective Function} &\multicolumn{2}{|c|}{Result}  \\
     \cline{1-3}
     \cline{4-5}
     ML &AT &VAT & Accuracy  &F1 Score\\
     \hline
     \checkmark & &  &0.910 &0.907 \\
     \hline
     \checkmark &\checkmark  & &0.936 &0.930 \\
     \hline
     &\checkmark &\checkmark &0.854 &0.860 \\
     \hline
     \checkmark & &\checkmark & 0.936 &0.930 \\
     \hline
     \checkmark &\checkmark &\checkmark & \textbf{0.954} &\textbf{0.953} \\
     \hline
    \end{tabular}}
\caption{Results of Cross-SEAN with different variations of the mixed objective function.}
\label{tab:ablitiona}
\vspace{-3mm}
\end{table}

\begin{table*}[h]
\centering
    \scalebox{0.75}{
    \begin{tabular}[H]{| c | c | c | c |c |c | c |c | c | c| c | c |}
     \hline
     \multicolumn{8}{|c|}{Linear Layers} &Attention &Cross-Stitch &\multicolumn{2}{c|}{Performance}  \\
     \cline{1-8}
     \cline{9-10}
     \cline{11-12}
    TF1 &UF1 &TF2 &UF2 &TF3 &UF3 &TF4 &UF4 & & &Accuracy &F1 Score \\
     \hline
    \multicolumn{2}{|c|}{128} & & & & & & & & &0.910 &0.884 \\
     \hline
    \multicolumn{2}{|c|}{64} &\multicolumn{2}{|c|}{256} & & & & &\checkmark & &0.932 &0.935 \\
     \hline
    64 &64 &256 &256 & & & & &\checkmark &\checkmark &0.931 &0.934 \\
     \hline
    \multicolumn{2}{|c|}{64} &\multicolumn{2}{|c|}{256} &\multicolumn{2}{|c|}{256} &\multicolumn{2}{|c|}{512} & & &0.939 &0.942 \\
     \hline
    \multicolumn{2}{|c|}{64} &\multicolumn{2}{|c|}{256} &\multicolumn{2}{|c|}{256} &\multicolumn{2}{|c|}{512} &\checkmark & &0.927 &0.944 \\
     \hline
    64 &64 &256 &256 & &256 & & &\checkmark &\checkmark &\textbf{0.954} &\textbf{0.953} \\
     \hline
    \end{tabular}}
\caption{Results with various fully-connected network combinations. Here, TF$i$ and UF$i$ represent the $i$th layer transposing a feature vector of tweet features and user features respectively. Two joined cells represent a concatenated form of the respective vectors feeding as inputs to the corresponding layer.}
\label{tab:ablitionb}
\vspace{-5mm}
\end{table*}

\subsection{Ablation Study}\label{sec:ablation}

\begin{figure}[htp]
  \centering
  \subfigure[ML Loss]{\includegraphics[width=0.225\linewidth]{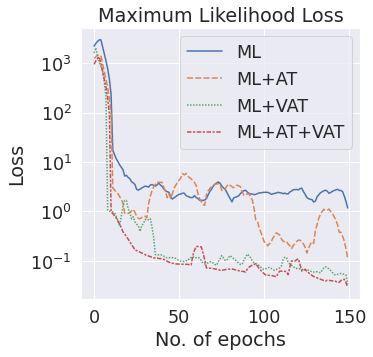}\label{fig:smoothML1.pdf}}\quad
  \subfigure[AT Loss]{\includegraphics[width=0.225\linewidth]{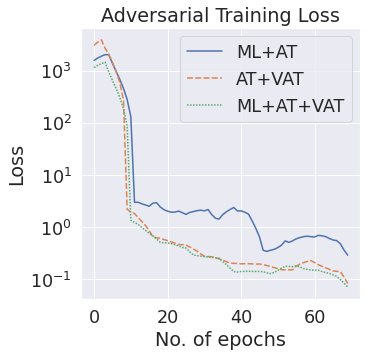}\label{fig:smoothAT.pdf}}\quad
  \subfigure[VAT Loss]{\includegraphics[width=0.225\linewidth]{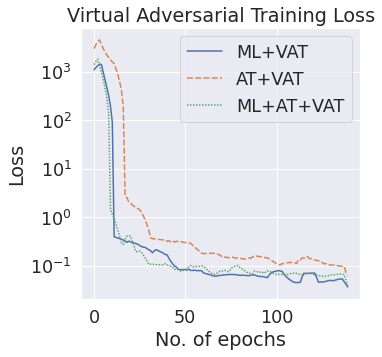}\label{fig:smoothVAT.pdf}}\quad
  \subfigure[Net Loss]{\includegraphics[width=0.225\linewidth]{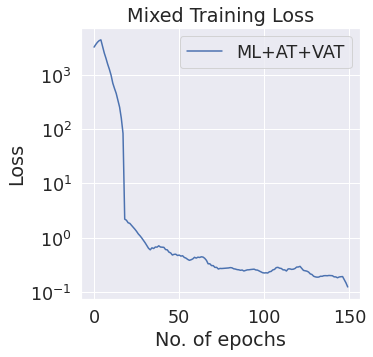}\label{fig:smoothnet.pdf}}\quad
  \caption{Variation of individual loss functions of Cross-SEAN with different combinations of the mixed objective function.}
  \label{fig:losses}
  \vspace{-5mm}
\end{figure}

\textbf{(a) Objective functions:} 
In Table \ref{tab:ablitiona}, we test the performance of Cross-SEAN on different combinations of the mixed objective function. We vary the values of $\lambda_{ML}$, $\lambda_{AT}$, $\lambda_{VAT}$ between $0$-$1$. A steady increase in the performance can be seen as we move from a vanilla supervised training objective (only maximum likelihood loss) to an additional semi-supervised mixed objective function.

Fig. \ref{fig:losses} shows the variation of different objectives functions -- ML, AT and VAT, individually, when trained with different combinations of the mixed objective function. For instance, Fig. \ref{fig:smoothML1.pdf} shows the variation of the individual ML Loss when different combinations of the net objective function is used.

From Fig. \ref{fig:smoothML1.pdf}, the regularisation effect of the two adversarial losses, AT and VAT, is apparent as it can be observed that their introduction considerably effects the individual ML loss, making it drop to a larger extent, in fewer iterations. Even though the introduction of AT alone seems to make the loss curve more stochastic, the net loss is considerably lower. This can be seen in addition to the surprising \textit{smoothing} effect which is observed wherever the VAT loss is considered, including Fig. \ref{fig:smoothAT.pdf} and \ref{fig:smoothVAT.pdf}. These two properties of AT and VAT losses respectively, motivate their usage together, thus resulting into an efficient and smooth decrease of loss and strengthening our hypothesis of leveraging unlabelled data. This is further ensured by another interesting observation by using only AT and VAT losses for the training -- although as expected, we achieve a deteriorated accuracy as shown in Table \ref{tab:ablitiona}, the corresponding losses in Figs. \ref{fig:smoothAT.pdf}-\ref{fig:smoothVAT.pdf} show high consistency and smoothness. Fig. \ref{fig:smoothnet.pdf} shows the final loss curve when all the 3 losses are used, i.e., when $\lambda_{ML}=\lambda_{AT}=\lambda_{VAT}=1$.

\noindent\textbf{(b) Model Components:} Fig. \ref{tab:ablitionb} shows the importance of different components used in Cross-SEAN such as cross-stitch, attention and feed-forward layers for tweet and user features. We experiment across several combinations of tweet features and user features with concatenation and usage of cross passing through various layers as shown in Fig. \ref{tab:ablitionb}. We find that the best architecture is with the cross-stitch on tweet and user features when one output of the cross-stitch is combined in the early stages of the network and the other output is fused in the later stage.
Also the use of attention shows performance improvement of the final model.

In our initial set of experiments, the cross-stitch was introduced between the encoded representation of the tweet text, obtained after passing it through Bi-LSTM, and a concatenated form of tweet and user features. A considerable difference in the performance is observed between the two, the former being the superior one. We relate this to the fact that the encoded representation of the tweet text is considerably different from the additional features, while they in themselves are very similar. Further, since the tweet features are inherently more similar to the tweet text, the cross-stitch output corresponding to the tweet features is first concatenated to the encoded tweet text and lastly with the user features. This is also shown in Table \ref{tab:ablitionb}, where the architecture used in the last row evidently outperforms the one in the 3rd row, which represents concatenation of the three outputs on the same level.

\section{Chrome-SEAN: A Chrome Extension \label{sec:chrome}}
Cross-SEAN is an end-to-end model which enables for identification of fake tweets in real time. Keeping the users warned is a very important step and would help with an easy access through the browser. In order to help users detect misinformation on Twitter in real time, we deploy Cross-SEAN as a Chrome browser extension, called Chrome-SEAN that replicates the performance of the model while performing a lot of other features as well.

Chrome-SEAN is built as Chrome extension, which uses jQuery\footnote{\url{https://jquery.com}} to send and receive requests from POST API method. We deployed the Cross-SEAN model using Flask\footnote{\url{https://flask.palletsprojects.com/en/1.1.x/}} in our local servers which can receive the POST API requests concurrently. To handle the load balancing over multiple concurrent requests, we use Redis\footnote{\url{https://redis.io/}}. The server is not burdened with resource intensive requests, and the combination of Flask and Redis performs efficient communication through APIs.

\begin{figure}
    \centering
    \includegraphics[width=0.8\textwidth]{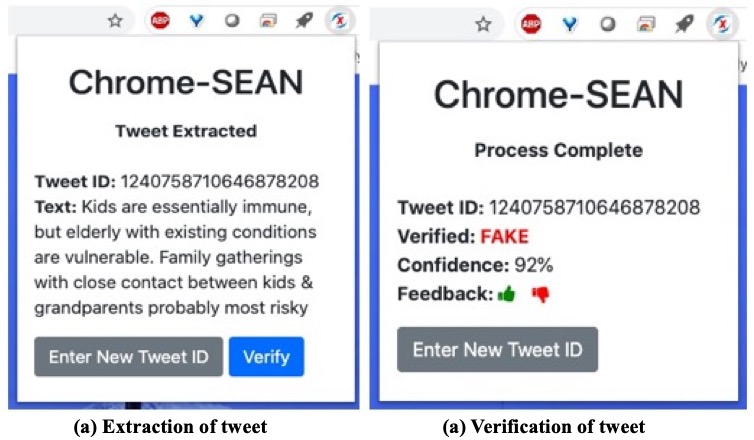}
    \caption{The working of Chrome-SEAN, a Chrome extension of Cross-SEAN.}
    \label{fig:extension}
    \vspace{-5mm}
\end{figure}

Chrome-SEAN first identifies the tweet ID through the URL while scanning Twitter, and sends it to the server using an API. Chrome-SEAN also provides the option to enter the tweet ID manually. Upon requesting to Cross-SEAN, the raw data is first transformed to the necessary format and then passed through the model. \textcolor{black}{The detected class along with its confidence from the softmax layer is returned back to the extension and displayed.} Fig. \ref{fig:extension} shows the working of Chrome-SEAN in two stages. \textcolor{black}{In the former stage, the extraction of the tweet is performed in the browser side and is instant, whereas in the latter stage, verification of the tweet takes on an average of 1.2 seconds per tweet (single API request).}

As shown in Fig. \ref{fig:extension}, we take users' feedback on our final classification output and consider it as a true label in the extended online dataset. Additionally, we employ an online training mechanism on the basis of users' feedback if it differs from the class identified and check the confidence of the model; the model is trained only if the confidence is lower than $0.6$. We take special care before online training to make the model robust to attackers attempting to pollute the results. To handle load balancing on the server, we make use of Redis.

\textbf{User Study:} Chrome-SEAN was tested by 35 users until now. We first randomly sampled tweets from the human-annotated set of tweets which were not a part of the training set, assigned them to users and asked them to test on similar tweets, totalling $215$ tweet inputs, ranging from a wide variety of sub-topics, users and timelines. It was observed that $67\%$ of these input tweets were made within the last $7$ days, $53\%$ were from new users with less than $5$ tweets, and $85\%$ had a retweets count of less than 10. 

We asked users to provide feedback on each tweet they tested with Chrome-SEAN, in accordance with the true label. We found that $203$ out of $215$ ratings were positive, i.e., deeming the prediction by Chrome-SEAN correct, resulting into an accuracy of $94\%$ and F1 Score of $94.3\%$. Such high level of accuracy on such a diverse set of inputs depicts Cross-SEAN's ability to pick the appropriate input features when making a prediction.

\section{Discussion and Conclusion \label{sec:concl}}

This work introduced the task of COVID-19 fake news detection on Twitter. We collected related tweets from diverse sources.
Post human annotations, we proposed \CTF, the first labelled Twitter dataset, consisting of COVID-19 related labelled genuine and fake tweets along with a huge set of unlabelled data. We also presented a thorough analysis to understand surface-level linguistic features.

As the amount of labelled data is limited, we made use of the vast unlabelled data to train the neural attention model in a semi-supervised fashion as learning the semantic structures of language around COVID-19 helps the model learn better. We collected external knowledge for all the tweets by taking the most relevant stance from credible sources on the web. 
As fake news around COVID-19 are emerging, even if the model is not trained on a certain fake news topics, we assume that external knowledge from a trusted source could help aid the classification. 
We built a neural attention model which takes various inputs such as tweet text, tweet features, user features and external knowledge for each tweet. We employed cross-stitch units for optimal sharing of parameters among tweet features and user features. As tweet text and tweet features are closely related, we performed optimal sharing of information by concatenating one output of cross-stitch early in the network and the other latter. 
Maximum likelihood and adversarial training are used for supervised loss, while virtual adversarial training for unsupervised loss. Usage of adversarial losses further adds regularisation and robustness to the model. We then incorporated this model into Cross-SEAN, a novel cross-stitch model which performs under a semi-supervised setting by leveraging both unlabelled and labelled data with optimal data sharing across various tweet information.

Cross-SEAN is highly effective, outperforming seven state-of-the-art models significantly. We contrasted features of baseline models with Cross-SEAN and showed various metrics. We showed a thorough ablation study with various fully-connected network combinations of the model and the respective accuracy contrasting the importance of individual components of the model. We also showed variation of individual loss functions with the different configurations of the mixed objective function.

To make use of Cross-SEAN in real time by general users, we developed Chrome-SEAN, a chrome extension based on Cross-SEAN to flag fake tweets, which showed reasonable performance in a small-scale user study. Chrome-SEAN is built to be robust to handle vast amount of concurrent requests. We introduced several features to Chrome-SEAN which can further help collect labelled data using user feedback. Cross-SEAN further trains in an online fashion, for a given feedback if the confidence of the model is low. Chrome-SEAN is further tested by human subjects.

\textbf{Shortcomings of Cross-SEAN:} We observe following shortcomings of Cross-SEAN:
\begin{itemize}
    \item The nature of language used in micro-blogging sites such as Twitter, in certain times makes the external knowledge noisy. Often times, a few trusted news sources on the Internet are biased on political topics which in turn create bias in the external knowledge.
    \item Although external knowledge adds additional information relative to the test time helping emerging fake news, it may not promise complete robustness and early detection.
    \item Although the tweet features, user features and external knowledge can attribute to general fake news, Cross-SEAN is a model specifically tuned for COVID-19 fake news, and is not tested on general fake news on Twitter.
\end{itemize}

\textbf{Future work:} 
We plan improve on the following points: 
\begin{itemize}
    \item We intend to study the dynamic graph structure of the follower-followee and tweet-retweet network, and extract representations from tweet and user nodes to help early detection of COVID-19 fake news.
    \item We will add additional improved filters to the process of extracting external knowledge to remove possible bias and noise.
    \item We will work towards explainability of Cross-SEAN using the current structures of attention mechanism.
    \textcolor{black}{\item We plan to incorporate semantic information from other forms of media such as images, GIFs or videos which are readily available with the tweets. Even the textual information present in such media will be extracted and used for detection.}

\end{itemize}

\section*{Acknowledgements}
The work was partly supported by the Accenture Faculty Award and MHRD (India) under the SPARC
programme project \#P620. T. Chakraborty would like to thank the generous support of the Ramanujan Fellowship (SERB) and Infosys Centre for AI, IIIT Delhi.

\bibliography{elsarticle-template}

\begin{thebibliography}{10}
\bibitem{fisher2013syrian}
M.~Fisher, Syrian hackers claim ap hack that tipped stock market by \$136
  billion. is it terrorism, Washington Post 23.
\end{thebibliography}

\end{document}